%% file: main.tex
\documentclass[sigconf]{acmart}
\AtBeginDocument{%
  }

\setcopyright{acmlicensed}
\copyrightyear{2018}
\acmYear{2018}
\acmDOI{XXXXXXX.XXXXXXX}
\acmConference[Conference acronym 'XX]{Make sure to enter the correct
  conference title from your rights confirmation email}{June 03--05,
  2018}{Woodstock, NY}
\acmISBN{978-1-4503-XXXX-X/2018/06}

\usepackage{amsmath}
\usepackage{booktabs}
\usepackage{multirow}
\usepackage{array}
\usepackage{needspace}
\usepackage[ruled,vlined,linesnumbered]{algorithm2e}
\usepackage{tikz}
\usepackage{hyphenat}
\usetikzlibrary{arrows.meta,positioning}
\begin{document}

\title{ElecLoad: Adaptive Electricity Load Forecasting Guided by Multiple Events}


\author{
Huajian Zhang$^{1}$,
Mingyue Cheng$^{1*}$,
Xingpeng Gao$^{1}$,
Xiaoyu Tao$^{1}$,
Zhiding Liu$^{1}$,
Jiqian Yang$^{2}$,
Zhi Li$^{2}$
}

\affiliation{
  \institution{$^{1}$State Key Laboratory of Cognitive Intelligence, University of Science and Technology of China}
  \city{Hefei}
  \country{China}
}

\affiliation{
  \institution{$^{2}$Shenzhen International Graduate School, Tsinghua University}
  \city{Shenzhen}
  \country{China}
}

\email{{zhjustc, gaoxingpeng, txytiny, zhiding, yangjq}@mail.ustc.edu.cn,  {mycheng}@ustc.edu.cn, {zhilizl}@sz.tsinghua.edu.cn}

\renewcommand{\shortauthors}{Huajian Zhang et al.}

\begin{abstract}
Accurate electricity load forecasting provides critical decision support for energy procurement, generation scheduling, and supply-demand balancing, contributing to the economic efficiency and operational reliability of power systems. However, future electricity demand can be affected by the complex interplay of multiple events described in Web notices, weather alerts, and event schedules. Their effects vary over time and can reinforce or offset one another, making electricity load difficult to forecast accurately. While existing methods can incorporate context as additional input, how to use the effects of multiple events and their interactions to guide the load forecasting process itself remains to be explored. To address this gap, we propose ElecLoad, a context-aware agent for multi-event electricity load forecasting that adaptively organizes the forecasting process. Specifically, we train ElecLoad to combine historical load with context to analyze each event's relevance to the target load, the timing of its effects, and potential interactions among events. Guided by this analysis, ElecLoad adaptively chooses to forecast the entire forecast horizon directly or to partition it into segments according to changes in event effects. When the horizon is partitioned, sub-agents independently generate initial forecasts for each segment. The main agent then reviews these forecasts, and the sub-agents perform selective refinement guided by its feedback. Experiments on multiple electricity load datasets show that ElecLoad achieves state-of-the-art performance in most evaluation scenarios. Code is available at \url{https://github.com/zhjai/ElecLoad}.
\vspace{-6pt}
\end{abstract}

\maketitle

\begin{figure}[t]
\centering
\includegraphics[width=\linewidth]{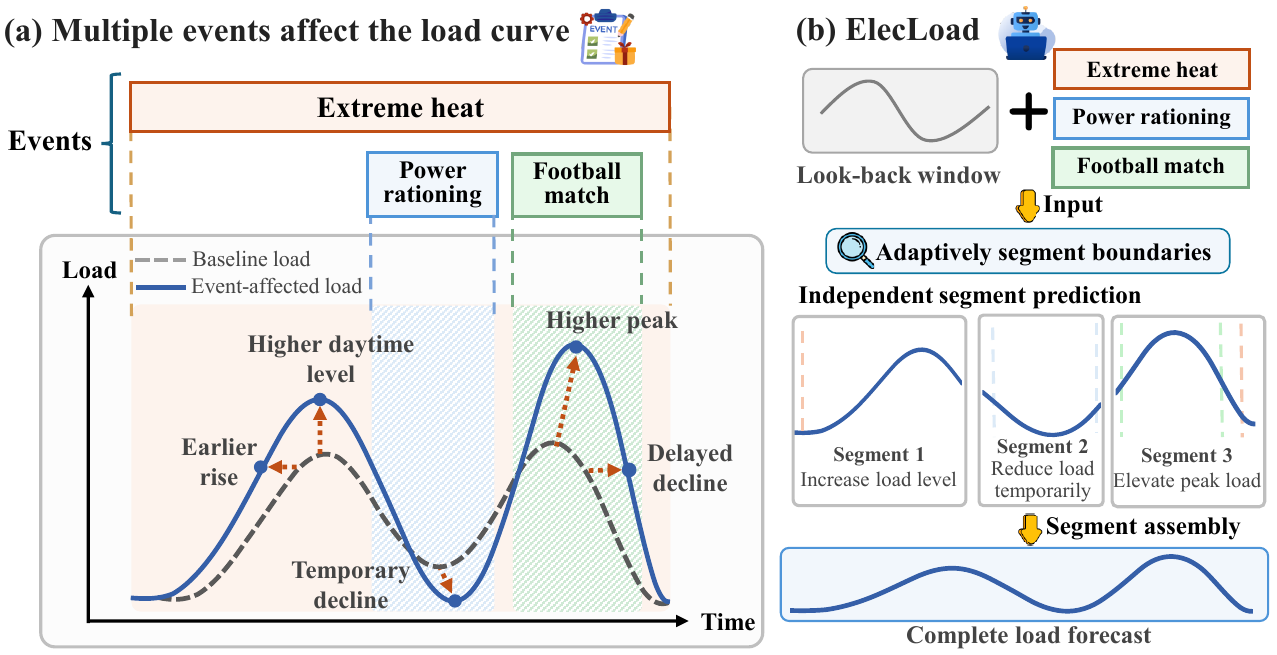}
\caption{Conceptual illustration of (a) load changes under multiple events over time and (b) event-guided forecast partitioning in ElecLoad.}
\label{fig:motivation}
\vspace{-6pt}
\end{figure}

\section{Introduction}

To meet residential and industrial electricity demand, power systems need to plan electricity procurement, generation scheduling, and reserve allocation in advance to maintain the balance between electricity supply and demand. These decisions depend on anticipating future electricity demand, and accurate electricity load forecasting can provide valuable support. Forecasting errors can have economic and operational consequences: overestimating demand may lead to unnecessary electricity procurement and resource commitments, whereas underestimating demand may increase the costs of unplanned adjustments and risks to system operation~\cite{wang2017improving}. Moreover, effective power system operation requires not only accurately forecasting future demand levels but also capturing the timing and magnitude of load peaks~\cite{shukla2024bigdeal}. Therefore, accurately forecasting future electricity load sequences is important for improving the economic efficiency and operational reliability of power systems. Reliable electricity supply also underpins the data centers on which Web services depend~\cite{iea2025energyai}.

However, extreme weather, major sporting events, and events related to power system operation can cause electricity load to deviate from regular consumption patterns. Weather alerts, news reports, and operational notices on the Web can provide clues about event details, including relevant times and regions~\cite{liu2026world2ts}. Using these Web event descriptions for forecasting requires assessing each event's relevance to the target load, the timing of its effects within the forecast horizon, and potential interactions among events. For example, during a heatwave in the United Kingdom on June 23, 2026, many viewers watched England's World Cup match with air conditioners and electric fans running. A report on The Guardian’s website cited estimates of additional electricity demand at halftime and after the final whistle~\cite{ambrose2026heatwave}. When event effects overlap or occur in sequence, they may reinforce or offset one another across different periods. Load may still be recovering from one event when another begins to affect consumption. At such times, load reflects the remaining effects of the earlier event alongside those of the later one. As illustrated in Figure~\ref{fig:motivation}(a), these changing influences can alter load levels and the timing and magnitude of load peaks. Analysis of changing event effects and load recovery must inform hourly numerical load forecasts and guide decisions about which periods to forecast together and which forecasts need refinement.

General forecasting methods already incorporate covariates or textual information~\cite{ansari2025chronos,ge2026eventtsf}, and some introduce process mechanisms such as event alignment, planning, and reflection~\cite{ge2026eventtsf,pan2026castflow,wang2024news}. In electricity load forecasting, MMGPT4LF~\cite{gao2025mmgpt4lf} aligns and fuses textual and time-series representations. ITA-LF~\cite{wang2026news} analyzes news relevance, effect direction, and duration before selecting forecasting inputs. NACF~\cite{chen2026counterfactual} represents news semantics and publication times together with historical load and covariates, and examines forecasts under altered event conditions. A remaining challenge is to use analysis of how the effects of multiple events change over time to guide an adaptive forecasting process.

To address this challenge, we propose ElecLoad, a context-aware agent for multi-event electricity load forecasting. ElecLoad combines historical load with context to analyze each event's relevance to the target load, the timing of its effects, and potential interactions among events. Based on this analysis, the main agent adaptively chooses between forecasting the full horizon directly and partitioning it into multiple segments. The partition follows changes in event effects over time, as shown in Figure~\ref{fig:motivation}(b). For segmented forecasting, ElecLoad adopts a multi-agent collaboration framework in which the main agent organizes the process and shares the same language model with the sub-agents. Each sub-agent independently generates an initial forecast using the complete original historical load and context, together with the main agent's analysis of its assigned segment. To address mismatches between predicted load changes and event effects, the main agent checks forecasts against the original inputs and event analysis, selecting segments for feedback and refinement. The corresponding sub-agent returns a refined forecast, and the main agent reassesses the updated sequence to decide whether to continue or end this selective refinement loop. Finally, boundary reconciliation retains or adjusts values near segment boundaries. Our main contributions are summarized as follows:
\begin{itemize}
\item We analyze the forecasting requirements arising from overlapping events, sequential event effects, and load recovery, and propose using changes in event effects over time to guide the forecasting process.
\item ElecLoad combines historical load and context to analyze each event's relevance to the target load, the timing of its effects, and potential interactions among events. This analysis informs numerical load predictions for different periods within the forecast horizon.
\item We evaluate ElecLoad against forecasting baselines on in-domain electricity load datasets and out-of-domain electricity systems, and analyze its performance under different event conditions.
\end{itemize}

\section{Related Work}

\subsection{Neural Time Series Forecasting}

{\emergencystretch=1em
One major line of research in neural time series forecasting improves how models represent the structure of a series~\cite{cheng2026comprehensive}. PatchTST~\cite{nie2022time} encodes local subsequences as patches for Transformer input, preserving local information while supporting longer input histories. ConvTimeNet~\cite{cheng2025convtimenet} uses deformable patches and hierarchical convolutions to capture local patterns and their multiscale dependencies. In representation learning, TimeDART~\cite{wang2025timedart} combines a causal Transformer with diffusion denoising in self-supervised pretraining to learn global and local representations for downstream forecasting. To account for differences in periodicity and trends across application domains, OneCast~\cite{pan2025onecast} decomposes a series into seasonal and trend components. It models these components separately and combines the resulting predictions to produce the final forecast. Beyond the structure of the series itself, external conditions can also inform forecasting: Chronos-2~\cite{ansari2025chronos} and TimesFM-3~\cite{jain2026timesfm} both support zero-shot forecasting using historical time series and covariates. EventTSF~\cite{ge2026eventtsf} aligns event embeddings with fixed-length segments and autoregressively predicts subsequent segments from preceding hidden states and event information.\par}

\vspace{-6pt}

\subsection{LLM-Based Forecasting}

Forecasting with LLMs connects contextual understanding and reasoning to numerical prediction. TokenCast discretizes time series into tokens, aligns them with context tokens, and decodes generated future tokens into numerical values~\cite{tao2026tokencast}. TimeReasoner performs multistep reasoning, using earlier outputs as context for subsequent forecasts in its rollout~\cite{cheng2026can}. MemCast builds hierarchical memory from training-set prediction results, reasoning trajectories, and temporal features to guide reasoning, trajectory selection, and reflective iteration during inference~\cite{tao2026memcast}. For evidence gathering and feedback, AlphaCast uses investigation, generation, and reflection roles to construct context, generate forecasts, and refine them~\cite{zhang2025alphacast}. Subsequently, CastFlow assigns planning and reflection to a frozen language model, while a fine-tuned model produces numerical forecasts from tool evidence and an ensemble forecast baseline~\cite{pan2026castflow}. CastFSR combines initial forecasts with contextual reasoning, adaptively retrieves evidence and selects a lookback window, and uses reflection to correct local inconsistencies~\cite{tao2026castfsr}. MoiraiAgent offers contextual forecasting and expert selection as independent, composable capabilities. The former adjusts historical windows and forecasts using context, while the latter selects suitable forecasting models~\cite{yan2026moiraiagent}.

\subsection{Electricity Load Forecasting}

Electricity load forecasting research explores how to incorporate event information. Gaussian-process transfer learning calibrates a prior forecast from a few observations during an anomalous event~\cite{zhao2023gaussian}. News and Load uses news features for day-ahead forecasting~\cite{bai2024news}; its follow-up combines news with economic indicators for regional load forecasting across multiple horizons~\cite{bai2025news}. MMGPT4LF encodes textual descriptions of calendar information, temperature, and historical load trends with GPT-2 and fuses them with load sequences through cross-attention~\cite{gao2025mmgpt4lf}. ITA-LF uses an LLM agent to analyze the relevance of news to load and the direction and duration of its impact, and to reflect on news selection. A fine-tuned LLM then forecasts from selected news and historical load~\cite{wang2026news}. NACF uses an event-conditioned network to produce factual forecasts from historical load, covariates, and the semantics and publication times of LLM-structured news events~\cite{chen2026counterfactual}. It also produces counterfactual forecasts under a no-news baseline or hypothetical events. AutoLFM uses LLM agents to automate the development of building load predictors and corrects generated code with validation feedback~\cite{li2026autolfm}. ElecLoad analyzes how the effects of multiple events on the target load vary over time and, guided by this analysis, adaptively chooses to forecast the full horizon directly or partition the forecast horizon into segments. On the segmented path, the event analysis and forecast checks determine which segments undergo selective refinement.

\section{Problem Formulation}

We study multi-event electricity load forecasting, covering both power system load and household electricity load. Let $t_0$ denote the forecast origin and $\Delta t$ the sampling interval. The load at time $t_0+\tau\Delta t$ is denoted by $y_\tau$, with $\tau=0$ marking the first time step to forecast. Given a historical load sequence of length $L$, $\mathbf y_{-L:-1}=(y_{-L},\ldots,y_{-1})$, the goal is to forecast the load sequence $\mathbf y_{0:H-1}=(y_0,\ldots,y_{H-1})$ over the next $H$ steps.

The context $\mathcal C=(\mathcal X,\mathcal E)$ consists of auxiliary inputs beyond the historical load sequence, with covariates $\mathcal X$ including numerical, categorical, and calendar features. Each event in the event set $\mathcal E=\{e_m\}_{m=1}^{M}$ is represented by a textual description and any temporal and geographic information supplied by its source. An announcement's publication time indicates when the information is released. The occurrence or effective period stated by the source provides cues for relating the event to the forecast horizon. Neither the publication time nor the stated period is equivalent to the time range over which the load is affected. The data protocol specifies the conditions under which future covariates are provided. In the experiments reported here, event announcements are available at the corresponding forecast origins.

A predictor $f$ maps the historical load sequence and the available context to a numerical load forecast:
\begin{equation}
\widehat{\mathbf y}_{0:H-1}
=
f\!\left(\mathbf y_{-L:-1},\,\mathcal C\right),
\end{equation}
where $\widehat{\mathbf y}_{0:H-1}\in\mathbb R^{H}$ is the predicted load sequence covering the entire forecast horizon.

\section{Method}

\begin{figure*}[t]
\centering
\includegraphics[width=0.92\textwidth]{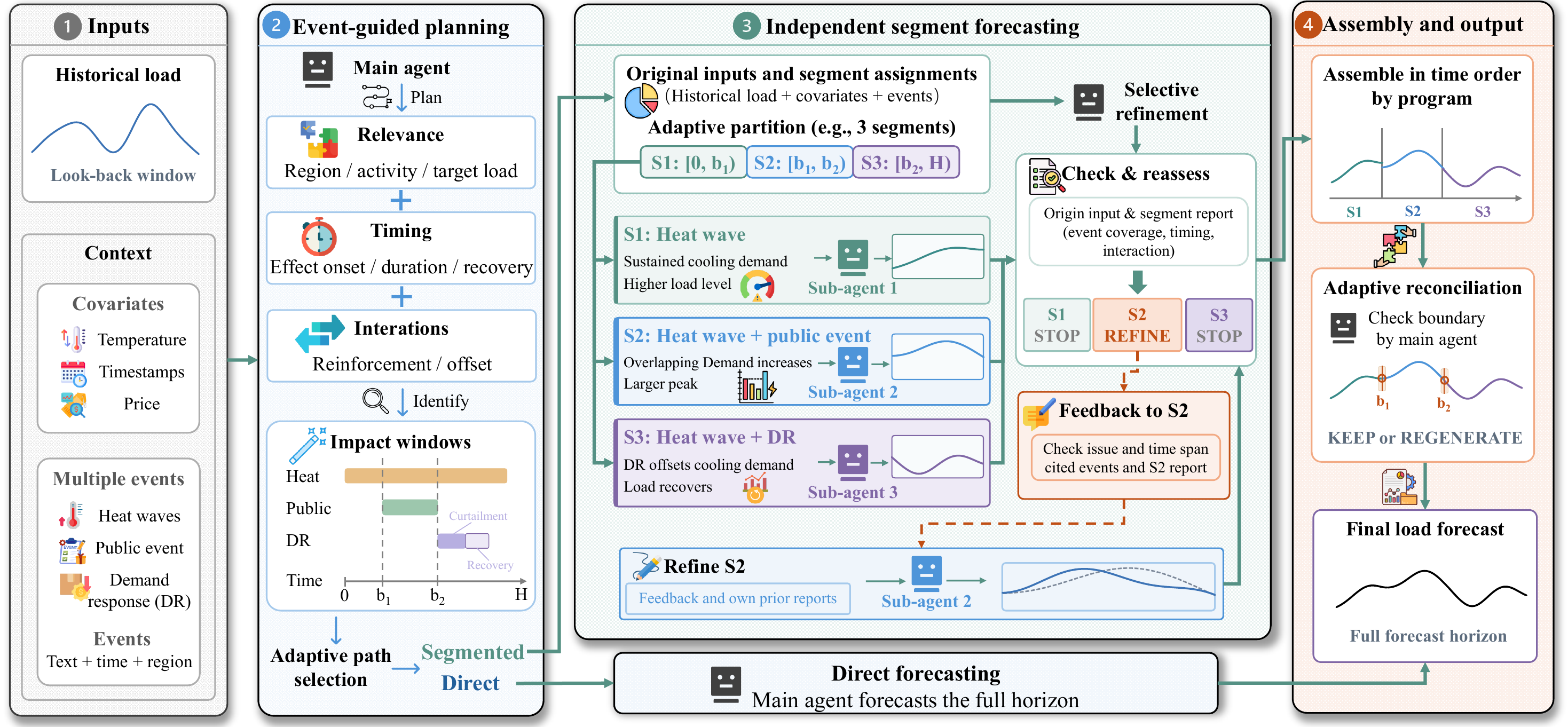}
\caption{ElecLoad uses multi-event analysis to adaptively choose direct or segmented forecasting. In the segmented path, the main agent selects segments for feedback and refinement. The main agent and sub-agents share the same LLM weights.}
\Description{Historical load and context enter ElecLoad. Event analysis guides the choice between direct forecasting of the full horizon and partitioning into segments. On the segmented path, sub-agents generate forecasts. The main agent provides feedback on selected segments; their sub-agents refine the forecasts and return them for reassessment. After stopping, segment forecasts are assembled in time order, with optional adjustments near segment boundaries. Both paths produce a final load forecast.}
\label{fig:method}
\end{figure*}

Figure~\ref{fig:method} and Algorithm~\ref{alg:elecload-workflow} show how ElecLoad uses multi-event analysis to guide adaptive forecasting. Its agents call four functions: event analysis and forecast planning (T1), numerical forecasting (T2), forecast checks with selective feedback and stopping decisions (T3), and boundary reconciliation (T4). In T1, the main agent analyzes how event effects vary over time and adaptively chooses between direct forecasting and partitioning the forecast horizon into segments. The direct path ends after T2 generates the forecast from the original inputs and T1's analysis. On the segmented path, sub-agents independently generate initial segment forecasts in T2. In T3, the main agent checks them and either stops or provides feedback on a selected segment, which its sub-agent refines in T2. After each refinement, the main agent reassesses the complete forecast in T3 and decides whether to continue within the refinement budget. T4 retains or adjusts values near segment boundaries as needed. All roles share LLM parameters and receive the complete original inputs $\mathbf x$ on every call. T1's analysis $\mathcal A$ comprises impact analysis, overall explanation, and uncertainty assessment. Segment-specific analysis, reports, and role-specific histories are abbreviated in the calls, with details in Sections~\ref{sec:segment} and~\ref{sec:refinement}.

\input{algorithms/workflow}

\subsection{Event-Guided Forecast Planning}
\label{sec:planning}

The main agent uses historical load and context to analyze each event's relevance to the target load, the timing of its effects, and potential interactions among events. Drawing on the activities, regions involved, and timing described in event descriptions, it explains the mechanisms through which an event may alter electricity demand. It infers the direction of each event's effects and its impact window, defined as the range of forecast steps during which load may be affected. The impact window may include delayed responses or recovery after the event ends. The main agent also examines the order of events and how their effects overlap. For example, it considers whether different effects may reinforce or offset one another, and whether another event may affect load recovery following an earlier event. This impact analysis draws on specific evidence in the inputs and retains uncertainty about the timing of effects and potential interactions.

Based on this impact analysis, the main agent selects one of two forecasting paths: direct forecasting or segmented forecasting. In direct forecasting, T2 generates the load forecast for the entire horizon from the original inputs and T1's analysis, and the process ends. On the segmented path, it partitions the forecast horizon into segments, inferring their number and boundaries from how event effects change over time. For example, a boundary can mark an effect's onset or gradual decay, the start of load recovery, or a change in how multiple effects overlap. One event's effects may span multiple segments, and one segment may contain the effects of multiple events. For a forecast horizon comprising $H$ discrete time steps, the partition is defined as
\begin{equation}
\begin{aligned}
\mathcal S &= \{S_j\}_{j=1}^{K},\\
S_j &= \{\tau\in\mathbb Z\mid b_{j-1}\leq\tau<b_j\},\\
0 &= b_0<b_1<\cdots<b_K=H,\quad K\geq2,
\end{aligned}
\end{equation}
where the integer boundaries $b_j$ yield chronological segments that cover all forecast steps without gaps or overlaps, within preset limits on segment count and minimum length.

\subsection{Independent Segment Forecasting}
\label{sec:segment}
On the segmented path, each sub-agent receives one forecast segment \(S_j\) for T2 forecasting. The program selects factors whose impact windows overlap \(S_j\) and passes their analysis and potential interactions. The sub-agent receives the complete original historical load and context, along with the main agent’s overall explanation and uncertainty assessment. For its initial forecast, each sub-agent receives no forecasts or reports generated by other sub-agents. This separation is intended to limit error propagation across segments during autoregressive generation.

The sub-agent evaluates the main agent's analysis against historical load and context, then predicts the load level, curve shape, and timing of changes within its segment. It translates the analyzed persistence, overlap, reinforcement, and offsetting into numerical values at each forecast step. Full original context supports effects spanning segment boundaries, such as recovery from an earlier event. All segments share the original forecast origin and future time grid, preserving each target step's lead time.

Each sub-agent returns a complete segment forecast and a report explaining its rationale and uncertainty, with a self-check against the evidence. We denote the complete initial forecast for segment \(S_j\) by \(\hat{\mathbf y}^{(0)}_{S_j}=(\hat y^{(0)}_\tau)_{\tau=b_{j-1}}^{b_j-1}\). Concatenating these segment forecasts in chronological order yields the initial forecast sequence covering the entire forecast horizon:
\begin{equation}
\hat{\mathbf y}^{(0)}
=
\operatorname{Concat}\!\left(
\hat{\mathbf y}^{(0)}_{S_1},\ldots,
\hat{\mathbf y}^{(0)}_{S_K}
\right)
=
\left(\hat y^{(0)}_0,\ldots,\hat y^{(0)}_{H-1}\right).
\label{eq:initial-forecast}
\end{equation}

\subsection{Selective Refinement}
\label{sec:refinement}
In T3, the main agent checks the initial forecast sequence \(\hat{\mathbf y}^{(0)}\) for consistency with the original inputs, impact analysis, and segment reports. It examines whether predicted levels and recovery reflect the reinforcement, offsetting, and persistence of event effects. Using these checks, the feedback history, and the remaining budget, it selects one segment for refinement or stops. Its diagnostic feedback identifies issues to examine and supporting evidence, optionally highlighting a time interval for closer attention.

In T2, the selected sub-agent reexamines its forecast using the inputs described in Section~\ref{sec:segment}, its own previous outputs, and the feedback. It explains whether the feedback warrants changes and returns a complete forecast and updated report for the same segment. The highlighted interval focuses review, while delayed effects or recovery may motivate changes elsewhere within that segment. Let the \(r\)-th refinement (\(r\geq1\)) select segment \(S_{j_r}\) and return the forecast \(\tilde{\mathbf y}^{(r)}_{S_{j_r}}\). The forecast sequence is updated as
\begin{equation}
\hat y^{(r)}_\tau=
\begin{cases}
\tilde y^{(r)}_\tau, & \tau\in S_{j_r},\\
\hat y^{(r-1)}_\tau, & \tau\notin S_{j_r},
\end{cases}
\qquad \tau=0,\ldots,H-1.
\label{eq:refinement-update}
\end{equation}
The main agent reassesses the complete forecast after each refinement, using all segment results and the refinement history. It stops when no refinement is selected or the budget is exhausted.

\textbf{Adaptive boundary reconciliation.} After refinement, the main agent reconciles forecast values near segment boundaries using the original inputs, planning output, final reports, and refinement history. It checks whether differences where segments join reflect event onset, decay, or recovery. T4 returns \texttt{KEEP} or a \texttt{REGENERATE} patch that adjusts values only near segment boundaries, keeping the partition fixed.

\subsection{Model Training}
\label{sec:training}

ElecLoad is built on Qwen3.5-4B~\cite{qwen2026qwen35} and is trained in two stages. Reasoning initialization teaches the model basic numerical load forecasting and event identification and understanding. Agent optimization then trains all roles to execute tasks and make decisions within the adaptive forecasting process. Each stage uses supervised fine-tuning (SFT) followed by reinforcement learning (RL), with all model parameters updated in both SFT and RL.

During reasoning initialization, the teacher predicts loads and explains them from historical load and context. It uses training labels to review and retain or rewrite the explanation to support the target answer. The accepted explanation and true load labels supervise the student.

Agent SFT starts from the reasoning RL checkpoint. Demonstrations are generated mainly by the teacher, starting with event analyses and plans. After label-assisted review, accepted plans are executed; demonstrations retain the role explanations, control decisions, and numerical forecasts generated and accepted. Each episode $e$ is a complete forecasting workflow execution. SFT averages valid response-token losses within episodes and across episodes.

Agent RL starts from the agent SFT checkpoint and uses utility
\begin{equation}
U(e)=-\frac{W(e)}{0.10}-0.01\frac{C(e)}{C_{\mathrm{ref}}},
\label{eq:reward}
\end{equation}
where $W(e)$ is fractional WAPE of the delivered forecast, including fallbacks, and $C(e)$ counts input and output tokens across all calls, including failed attempts. The fixed per-case reference $C_{\mathrm{ref}}$ normalizes cost; $0.10$ sets the error scale and $0.01$ is the cost weight. We sample two independent planning responses per case and continue each accepted plan along both forecasting paths. Paired utilities support path selection. Role-specific credit supplies learning signals for initial forecasting, refinement decisions, and boundary reconciliation. It distinguishes local forecast quality or improvement, call cost, and output format and structural validity. Appendix~\ref{app:runtime} details supervision, credit assignment, and validation scoring.

\section{Experiments}

\subsection{Experimental Setup}

\subsubsection{Datasets}

We evaluate hourly household, regional, and system load, treating each power system as a domain. In-domain (ID) sources comprise the Global Energy Forecasting Competition 2014 (GEFCom2014) load track~\cite{hong2016probabilistic}, national load (Nord)~\cite{entsoePowerStats}, PJM native load and balancing-area demand~\cite{pjmDataMiner,eiaGridMonitor}, Southwest Power Pool (SPP) balancing-area demand~\cite{eiaGridMonitor}, Low Carbon London (LCL)~\cite{lclSmartMeter}, and iFlex~\cite{iflexDataset}. The Australian Energy Market Operator (AEMO)~\cite{aemoNemweb} and Electric Reliability Council of Texas (ERCOT)~\cite{ercotHourlyLoad} are out-of-domain (OOD) systems reserved for 96-hour multi-day and 24-hour day-ahead testing, respectively. Nord, PJM, and SPP use both 24- and 96-hour forecast horizons, while GEFCom2014 uses only the 96-hour horizon. We use LCL and iFlex only for day-ahead forecasting because the LCL tariff notices and iFlex research price schedules used here cover only the next day. Table~\ref{tab:data} lists sources, history and forecast windows, and split sizes.

\begin{table}[!t]
 \caption{Dataset sizes by forecasting window. $L/H$: history/forecast hours. ERCOT includes 71 system-level cases for the main evaluation and 14 regional sensitivity cases.}
 \label{tab:data}
 \centering\fontsize{6.5}{7.8}\selectfont
 \setlength{\tabcolsep}{1pt}
 \renewcommand{\arraystretch}{1.12}
 \begin{tabular*}{\columnwidth}{@{\extracolsep{\fill}}llrrrrr@{}}
  \toprule
  Source & $L/H$ & Train & Val. & ID test & OOD test & Total \\
  \midrule
  GEFCom2014 & 96/96 & 640 & 100 & 200 & 0 & 940 \\
  \multirow{2}{*}{Nord} & 168/24 & 288 & 48 & 96 & 0 & 432 \\
   & 96/96 & 72 & 12 & 24 & 0 & 108 \\
  \multirow{2}{*}{PJM} & 168/24 & 320 & 56 & 112 & 0 & 488 \\
   & 96/96 & 80 & 14 & 28 & 0 & 122 \\
  \multirow{2}{*}{SPP} & 168/24 & 160 & 24 & 48 & 0 & 232 \\
   & 96/96 & 40 & 6 & 12 & 0 & 58 \\
  LCL & 168/24 & 120 & 24 & 48 & 0 & 192 \\
  iFlex & 168/24 & 80 & 16 & 32 & 0 & 128 \\
  \midrule
  AEMO & 96/96 & 0 & 0 & 0 & 150 & 150 \\
  ERCOT & 168/24 & 0 & 0 & 0 & 85 & 85 \\
  \midrule
  Total & --- & 1,800 & 300 & 600 & 235 & 2,935 \\
  \bottomrule
 \end{tabular*}
\end{table}

To evaluate forecasting under different event conditions, we construct synthetic cases from GEFCom2014 with single or multiple load-relevant events, or only irrelevant events. We also retain the original load sequences without added events. For load-relevant events, we modify the future target load according to the specified magnitudes and timing of their effects. Some events include a recovery phase in which deferred electricity demand returns in later hours. We combine multiple events by summing effects computed separately from the original sequence or applying events sequentially by start time to the already modified sequence. Under the sequential rule, later effects depend on earlier changes. Cases with only irrelevant events retain the original future target load. Appendix~\ref{app:controlled} details load and event sources, construction rules, and case proportions.

\subsubsection{Evaluation Metrics}

We use weighted absolute percentage error (WAPE)~\cite{kolassa2007advantages} as the primary metric. The secondary metric is normalized root mean squared error (NRMSE)~\cite{koponen2020assessing}, which is more sensitive to large deviations. For a true load sequence $y$ and a forecast sequence $\hat y$, both of length $H$, the metrics are defined as

\begin{align}
\mathrm{WAPE}&=\frac{\sum_{k=0}^{H-1}|\hat y_k-y_k|}{\sum_{k=0}^{H-1}|y_k|},\\
\mathrm{NRMSE}&=\frac{\sqrt{H^{-1}\sum_{k=0}^{H-1}(\hat y_k-y_k)^2}}{H^{-1}\sum_{k=0}^{H-1}|y_k|}.
\label{eq:metrics}
\end{align}

WAPE and NRMSE score the delivered forecasts for all cases, including fallback forecasts after failure, and are reported as percentages. Appendix~\ref{app:aggregation} details aggregation and near-zero-load handling; Appendix~\ref{app:runtime-settings} describes fallback.

\begin{table*}[!t]
 \caption{Overall performance in multi-day and day-ahead load forecasting. WAPE and NRMSE are reported as percentages. Best results are bold; second-best results are underlined. $L/H$: history length/forecast horizon in hours.}
 \label{tab:matched-main}
 \centering\fontsize{6.5}{7.8}\selectfont
 \setlength{\tabcolsep}{0.65pt}
 \renewcommand{\arraystretch}{1.18}
 \begin{tabular*}{\textwidth}{@{\extracolsep{\fill}}l*{22}{c}@{}}
  \toprule
  & \multicolumn{10}{c}{Multi-day forecasting ($L=96,\ H=96$)} & \multicolumn{12}{c}{Day-ahead forecasting ($L=168,\ H=24$)} \\
  \cmidrule(lr){2-11} \cmidrule(lr){12-23}
  \multicolumn{1}{l}{\raisebox{-4pt}[0pt][0pt]{Method}} & \multicolumn{2}{c}{GEFCom2014} & \multicolumn{2}{c}{Nord} & \multicolumn{2}{c}{PJM} & \multicolumn{2}{c}{SPP} & \multicolumn{2}{c}{AEMO$^{\dagger}$} & \multicolumn{2}{c}{Nord} & \multicolumn{2}{c}{PJM} & \multicolumn{2}{c}{SPP} & \multicolumn{2}{c}{LCL} & \multicolumn{2}{c}{iFlex} & \multicolumn{2}{c}{ERCOT$^{\dagger}$} \\
  \cmidrule(lr){2-3} \cmidrule(lr){4-5} \cmidrule(lr){6-7} \cmidrule(lr){8-9} \cmidrule(lr){10-11} \cmidrule(lr){12-13} \cmidrule(lr){14-15} \cmidrule(lr){16-17} \cmidrule(lr){18-19} \cmidrule(lr){20-21} \cmidrule(lr){22-23}
  & WAPE & NRMSE & WAPE & NRMSE & WAPE & NRMSE & WAPE & NRMSE & WAPE & NRMSE & WAPE & NRMSE & WAPE & NRMSE & WAPE & NRMSE & WAPE & NRMSE & WAPE & NRMSE & WAPE & NRMSE \\
  \midrule
  ARIMAX & 17.63 & 20.91 & 7.49 & 8.87 & 10.09 & 11.62 & 6.09 & 7.08 & 29.21 & 33.83 & 5.48 & 6.58 & 7.01 & 8.07 & 6.25 & 7.11 & 808.45 & 964.88 & 30.19 & 37.34 & 7.44 & 8.86 \\
  LightGBM & 9.03 & 10.90 & 5.93 & 6.90 & 7.07 & 8.15 & 4.50 & 5.51 & 15.57 & 20.06 & 4.84 & 5.56 & 4.89 & 5.75 & 4.76 & 5.64 & 1008.97 & 1167.11 & 34.94 & 41.98 & 7.55 & 8.51 \\
  PatchTST & 13.82 & 16.91 & 7.97 & 9.35 & 7.52 & 8.96 & 5.07 & 6.23 & 18.45 & 22.67 & 5.22 & 6.14 & 5.08 & 6.28 & 4.33 & 5.46 & 1873.00 & 2151.25 & 31.19 & 39.03 & 7.00 & 8.56 \\
  TimeXer & 13.40 & 16.45 & 8.18 & 9.58 & 7.84 & 9.21 & 4.92 & 5.96 & 15.85 & 20.11 & 4.96 & 5.69 & 4.71 & 5.74 & 4.16 & 5.01 & 958.83 & 1120.30 & 30.49 & 38.58 & 7.18 & 8.47 \\
  EventTSF & 16.57 & 20.15 & 8.82 & 10.15 & 9.69 & 11.67 & 7.61 & 9.43 & 23.62 & 29.53 & 5.54 & 6.35 & 5.48 & 6.55 & 5.73 & 6.73 & 804.16 & 1017.90 & 27.65 & 35.52 & 11.77 & 13.36 \\
  Chronos-2 & 9.92 & 12.29 & 7.11 & 8.22 & 6.90 & 8.33 & 4.30 & 5.27 & 13.99 & 18.38 & 3.60 & 4.22 & 3.61 & 4.40 & \underline{2.64} & \underline{3.18} & 62.39 & 99.54 & \underline{25.39} & \underline{32.73} & \underline{4.28} & 5.23 \\
  TimesFM-3 & 9.40 & 11.68 & 7.13 & 8.45 & \underline{6.11} & \underline{7.48} & 4.23 & 5.12 & 14.75 & 19.26 & \textbf{3.29} & \textbf{3.91} & \underline{3.43} & \underline{4.23} & 2.66 & 3.28 & \textbf{41.87} & \textbf{73.86} & 25.64 & 33.02 & \textbf{3.99} & \textbf{4.97} \\
  TimeReasoner & 8.61 & 10.84 & 5.76 & \underline{6.76} & 6.45 & 7.78 & 3.82 & 4.87 & \textbf{12.61} & \textbf{17.48} & 3.96 & 4.55 & 5.48 & 6.30 & 2.94 & 3.52 & 97.70 & 179.39 & 30.10 & 38.72 & 12.94 & 13.78 \\
  MemCast & \underline{8.18} & \underline{10.35} & \underline{5.75} & 6.78 & 6.32 & 7.63 & \textbf{3.59} & \underline{4.53} & \underline{12.78} & \underline{17.57} & 4.52 & 5.26 & 3.89 & 4.76 & 5.75 & 6.98 & 67.93 & 142.36 & 31.16 & 40.31 & 7.85 & 9.53 \\
  AlphaCast & 12.68 & 15.39 & 7.93 & 9.22 & 7.61 & 9.03 & 4.51 & 5.51 & 14.20 & 18.73 & 3.96 & 4.57 & 3.64 & 4.45 & 2.93 & 3.48 & 2726.63 & 2890.32 & 30.53 & 38.33 & 5.02 & 6.04 \\
  CastFSR & 9.36 & 11.66 & 10.66 & 11.74 & 6.66 & 7.95 & 3.82 & 4.80 & 13.39 & 17.69 & 3.61 & 4.15 & 3.78 & 4.58 & 3.33 & 3.87 & 86.95 & 167.05 & 26.88 & 34.78 & 5.73 & 6.66 \\
  \midrule
  MMGPT4LF & 10.86 & 13.25 & 7.71 & 8.88 & 8.80 & 10.11 & 6.84 & 7.98 & 18.44 & 23.23 & 7.00 & 7.91 & 7.72 & 9.04 & 7.60 & 8.80 & 851.35 & 938.90 & 30.62 & 38.23 & 13.75 & 15.46 \\
  NACF & 14.23 & 17.15 & 7.88 & 9.22 & 8.81 & 10.41 & 6.69 & 8.07 & 17.22 & 21.73 & 5.96 & 7.06 & 5.92 & 7.17 & 5.14 & 6.42 & 1761.87 & 2013.86 & 33.87 & 42.33 & 11.19 & 13.29 \\
  ITA-LF & 15.94 & 19.66 & 9.69 & 11.55 & 14.63 & 22.94 & 9.55 & 11.21 & 32.68 & 48.16 & 5.06 & 7.18 & 4.18 & 5.13 & 4.04 & 4.83 & 65.19 & 107.48 & 32.66 & 43.73 & 6.37 & 7.62 \\
  \midrule
  \textbf{ElecLoad} & \textbf{6.68} & \textbf{8.42} & \textbf{5.43} & \textbf{6.43} & \textbf{6.03} & \textbf{7.28} & \underline{3.63} & \textbf{4.13} & 13.49 & 17.69 & \underline{3.52} & \underline{4.14} & \textbf{3.13} & \textbf{3.86} & \textbf{2.36} & \textbf{2.92} & \underline{41.94} & \underline{77.24} & \textbf{24.32} & \textbf{31.06} & 4.33 & \underline{5.20} \\
  \bottomrule
 \end{tabular*}
 \par\vspace{2pt}
 {\raggedright $^{\dagger}$ OOD datasets; all other datasets are ID.\par}
\end{table*}

\subsubsection{Baselines}

The statistical baseline is ARIMAX; the tree-based machine learning baseline is LightGBM~\cite{ke2017lightgbm}; deep time series baselines include PatchTST~\cite{nie2022time}, TimeXer~\cite{wang2024timexer}, and EventTSF~\cite{ge2026eventtsf}. Time series foundation models include TimesFM-3~\cite{jain2026timesfm} and Chronos-2~\cite{ansari2025chronos}. Large language model forecasting baselines include TimeReasoner~\cite{cheng2026can} and MemCast~\cite{tao2026memcast}. We use a DeepSeek-v4-flash~\cite{deepseek2026v4flash} implementation of TimeReasoner in our experiments. Agent-based forecasting baselines include CastFSR~\cite{tao2026castfsr} and AlphaCast~\cite{zhang2025alphacast}.

Electricity load forecasting baselines include MMGPT4LF~\cite{gao2025mmgpt4lf}, NACF~\cite{chen2026counterfactual}, and ITA-LF~\cite{wang2026news}. Appendix~\ref{app:baseline-implementations} describes these variants.

\subsubsection{Implementation Details}

ElecLoad uses Qwen3.5-4B as its backbone and is trained on four NVIDIA H100 80GB GPUs. Agent SFT starts from the reasoning RL checkpoint and updates all model parameters for 2 epochs. It uses 11,135 role-specific training examples from 1,800 cases, with a batch size of 16. We use AdamW~\cite{loshchilov2017decoupled} with a learning rate of $5\times10^{-6}$ and a weight decay of 0.01. Agent RL starts from the resulting SFT checkpoint and also updates all model parameters. It uses a learning rate of $1\times10^{-6}$, 16 forecast origins per batch, and a sampling temperature of 1.0. For segmented forecasting, the implementation allows at most 4 segments, each containing at least 6 hourly steps. Each episode has a budget of 2 refinements, each applied to one selected segment. The main agent can stop before using the full budget. Appendix~\ref{app:runtime-settings} lists training configurations and runtime budgets.

\subsection{Overall Performance}
\label{sec:overall-performance}

Table~\ref{tab:matched-main} compares ElecLoad with 14 external baselines across 22 metric comparisons. ElecLoad ranks first in 13 and second in six, with a top-two rank in all 18 in-domain comparisons. These ID results span 24-hour day-ahead and 96-hour multi-day forecasting and household, regional, and system load sources. On GEFCom2014, ElecLoad reduces WAPE and NRMSE by 18.3\% and 18.6\%, respectively, relative to the strongest baseline, MemCast.

AEMO and ERCOT test transfer to power systems excluded from task training, at forecast horizons of 96 and 24 hours, respectively. ElecLoad ranks second to fourth across these four OOD metric comparisons among the 15 evaluated methods.\vspace{-6pt}
\begin{figure*}[t]
 \centering
 \includegraphics[width=\textwidth]{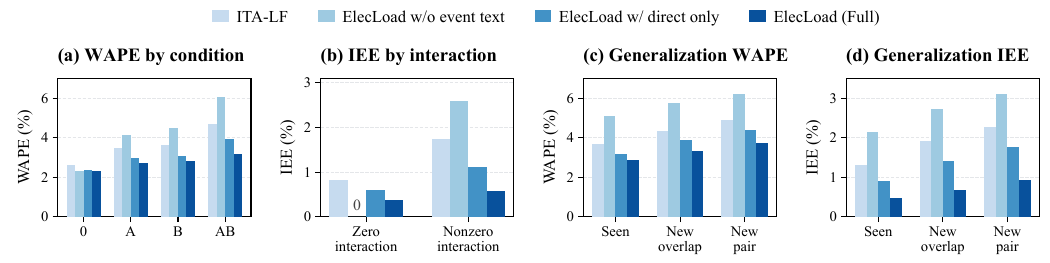}
 \caption{Forecasting on GEFCom2014 synthetic event cases. (a) WAPE for four event conditions; $0$ retains the original target without added events. (b) IEE grouped by zero/nonzero target interaction. (c,d) WAPE and IEE across generalization settings.}
 \label{fig:multi-event-interaction-demo}
 \label{fig:multi-event-generalization-demo}
\end{figure*}

\subsection{Ablation Study}
\label{sec:ablation}

We use the comparisons in Table~\ref{tab:ablation-demo} to examine how event text and covariates, together with planning, selective refinement, and boundary reconciliation, contribute to ElecLoad's forecasting accuracy. ElecLoad has the lowest WAPE and NRMSE on all three datasets. Removing event text or covariates raises both errors throughout. Covariate removal has the larger effect on PJM.

The segmented-only variant outperforms the direct-only variant on GEFCom2014, whereas the latter performs better on LCL; ElecLoad has lower mean errors than both. Equal-length segments also raise errors while preserving each case's segment count, highlighting the contribution of segment placement in planning (T1). Removing T3-guided refinement raises both metrics throughout. Refinement without feedback and random refinement also yield higher errors. Boundary reconciliation (T4) adds a smaller improvement: removing it, or replacing the option to retain the current sequence with forced boundary reconciliation within permitted limits, raises both metrics on every dataset.

\begin{table}[!t]
 \caption{Ablation study ($L=96$, $H=96$ for GEFCom2014; $L=168$, $H=24$ for PJM and LCL).}
 \label{tab:ablation-demo}
 \centering\fontsize{6.5}{7.8}\selectfont
 \setlength{\tabcolsep}{1pt}
 \renewcommand{\arraystretch}{1.12}
 \begin{tabular*}{\columnwidth}{@{\extracolsep{\fill}}ll*{6}{c}@{}}
  \toprule
  \multirow{2}{*}{Group} & \multirow{2}{*}{Variant} & \multicolumn{2}{c}{GEFCom2014} & \multicolumn{2}{c}{PJM} & \multicolumn{2}{c}{LCL} \\
  \cmidrule(lr){3-4} \cmidrule(lr){5-6} \cmidrule(lr){7-8}
  & & WAPE & NRMSE & WAPE & NRMSE & WAPE & NRMSE \\
  \midrule
  Reference & \textbf{ElecLoad (Full)} & \textbf{6.68} & \textbf{8.42} & \textbf{3.13} & \textbf{3.86} & \textbf{41.94} & \textbf{77.24} \\
  \midrule
  \multirow{2}{*}{Inputs} & w/o event text & 8.46 & 10.63 & 3.51 & 4.37 & 46.27 & 83.16 \\
   & w/o covariates & 7.39 & 9.36 & 3.79 & 4.71 & 44.12 & 80.58 \\
  \midrule
  \multirow{3}{*}{Planning} & w/ direct only & 7.82 & 9.91 & 3.48 & 4.32 & 42.91 & 78.83 \\
   & w/ segmented only & 6.97 & 8.86 & 3.39 & 4.21 & 45.36 & 82.07 \\
   & w/ equal-length segments & 7.51 & 9.54 & 3.46 & 4.29 & 43.68 & 80.01 \\
  \midrule
  \multirow{3}{*}{Refinement} & w/o refinement & 7.24 & 9.18 & 3.37 & 4.18 & 44.63 & 81.62 \\
   & w/o feedback & 7.06 & 8.97 & 3.29 & 4.08 & 43.57 & 79.85 \\
   & w/ random refinement & 6.94 & 8.81 & 3.32 & 4.12 & 43.06 & 79.18 \\
  \midrule
  \multirow{2}{*}{Reconciliation} & w/o reconciliation & 6.77 & 8.56 & 3.15 & 3.90 & 42.08 & 77.51 \\
   & w/ forced reconciliation & 6.73 & 8.51 & 3.19 & 3.97 & 42.31 & 77.96 \\
  \bottomrule
 \end{tabular*}
 \par\vspace{3pt}
 {\raggedright Removing feedback retains refinement; random refinement selects refinement targets at random.\par}
 \par\vspace{2pt}

\par\vspace{6pt}
 \normalsize
 \caption{Forecasting performance under different event conditions ($L=96$, $H=96$ for GEFCom2014; $L=168$, $H=24$ for PJM and LCL).}
 \label{tab:event-strata-demo}
 \centering\fontsize{6.5}{7.8}\selectfont
 \setlength{\tabcolsep}{1pt}
 \renewcommand{\arraystretch}{1.12}
 \begin{tabular*}{\columnwidth}{@{\extracolsep{\fill}}p{39pt}l*{6}{c}@{}}
  \toprule
  \multirow{2}{39pt}{\shortstack[l]{Event\\Condition}} & \multirow{2}{*}{Method} & \multicolumn{2}{c}{GEFCom2014} & \multicolumn{2}{c}{PJM} & \multicolumn{2}{c}{LCL} \\
  \cmidrule(lr){3-4} \cmidrule(lr){5-6} \cmidrule(lr){7-8}
  & & WAPE & NRMSE & WAPE & NRMSE & WAPE & NRMSE \\
  \midrule
 \multirow{4}{39pt}{No Event} & ITA-LF & 13.50 & 16.50 & 3.34 & 3.80 & 56.08 & 89.68 \\
  & ElecLoad w/o event text & \textbf{5.44} & \textbf{6.83} & \textbf{2.41} & \textbf{2.88} & 39.96 & 69.54 \\
  & ElecLoad w/ direct only & 6.00 & 7.60 & 2.80 & 3.40 & 38.72 & \textbf{69.47} \\
   & ElecLoad (Full) & \textbf{5.44} & \textbf{6.83} & 2.63 & 3.19 & \textbf{38.59} & 70.28 \\
  \midrule
 \multirow{4}{39pt}{\shortstack[l]{Single Event}} & ITA-LF & 15.80 & 19.40 & 4.12 & 5.10 & 64.39 & 103.40 \\
  & ElecLoad w/o event text & 8.20 & 10.30 & 3.57 & 4.43 & 47.64 & 84.39 \\
  & ElecLoad w/ direct only & 7.90 & 10.00 & 3.42 & 4.25 & 42.68 & 77.34 \\
   & ElecLoad (Full) & \textbf{6.42} & \textbf{8.06} & \textbf{3.12} & \textbf{3.84} & \textbf{41.73} & \textbf{76.38} \\
  \midrule
 \multirow{4}{39pt}{\shortstack[l]{Multiple\\Events}} & ITA-LF & 19.00 & 23.50 & 5.18 & 6.63 & 81.73 & 142.29 \\
  & ElecLoad w/o event text & 10.80 & 13.80 & 4.63 & 5.91 & 57.08 & 107.93 \\
  & ElecLoad w/ direct only & 9.50 & 12.10 & 4.31 & 5.43 & 51.00 & 97.68 \\
   & ElecLoad (Full) & \textbf{7.61} & \textbf{9.71} & \textbf{3.69} & \textbf{4.62} & \textbf{48.44} & \textbf{90.84} \\
  \midrule
 \multirow{4}{39pt}{\shortstack[l]{Irrelevant\\Events}} & ITA-LF & 14.00 & 17.20 & 4.39 & 5.48 & 58.56 & 94.55 \\
  & ElecLoad w/o event text & 8.02 & 9.08 & 3.83 & 4.81 & 40.40 & \textbf{70.78} \\
  & ElecLoad w/ direct only & 6.90 & 8.60 & 3.64 & 4.53 & 39.24 & 70.83 \\
   & ElecLoad (Full) & \textbf{5.49} & \textbf{6.91} & \textbf{3.26} & \textbf{4.04} & \textbf{39.00} & 71.46 \\
  \bottomrule
 \end{tabular*}
\end{table}

\begin{figure}[!t]
 \centering
 \includegraphics[width=\columnwidth]{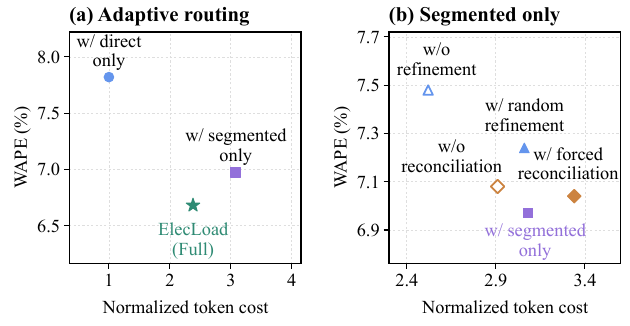}
 \caption{WAPE versus normalized token cost on GEFCom2014: (a) routing; (b) refinement and reconciliation.}
 \label{fig:planning-accuracy-cost-demo}
 \Description{Two square scatter plots show WAPE (\%) vertically against normalized token cost horizontally for GEFCom2014 policies. Adaptive routing compares direct-only, segmented-only, and ElecLoad (Full). Segmented only compares segmented-only, no refinement, random refinement, no reconciliation, and forced reconciliation on the same cases under forced segmented forecasting. Points are unconnected, and segmented-only has identical coordinates in both panels.}
\end{figure}

\begin{figure*}[!t]
 \centering
 \includegraphics[width=0.92\textwidth]{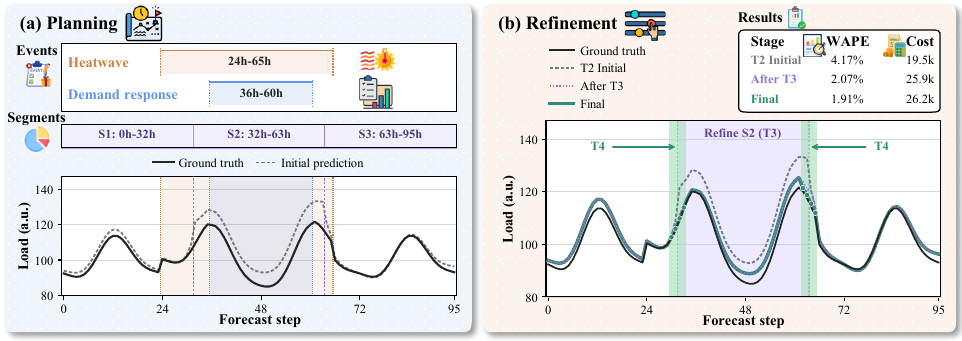}
 \caption{ElecLoad case: (a) T1 planning and T2 initial forecasts; (b) T3-guided refinement and T4 boundary reconciliation.}
 \label{fig:planning-case-demo}
 \Description{Two panels show a 96-step electricity load forecast, with forecast step on the horizontal axes and load in arbitrary units on the vertical axes. Panel a shows a heatwave from steps 24 to 65 and demand response from steps 36 to 60 above three event-guided segments, S1, S2, and S3. The lower plot compares Ground truth with Initial prediction, which is too high within S2. Panel b compares Ground truth, Initial, After T3, and Final. Purple shading identifies S2 for refinement, and green bands with T4 arrows identify reconciliation near its boundaries. The summary reports WAPE values of 4.17\%, 2.07\%, and 1.91\% and cumulative costs of 19.5k, 25.9k, and 26.2k tokens for Initial, After T3, and Final, respectively.}
\end{figure*}

\subsection{Multi-Event Forecasting}
\label{sec:multi-event}
\textbf{Event conditions.} Table~\ref{tab:event-strata-demo} groups cases by event relevance. Relevant events affect the target load within the forecast horizon, including recovery. Irrelevant descriptions concern events elsewhere that do not affect this target, administrative notices unrelated to load, or events whose effects fall entirely outside the horizon. ElecLoad outperforms ITA-LF across groups; removing event text while retaining load history and covariates increases error when relevant events are present. Its advantage over the direct-only variant is larger for multiple events: PJM WAPE is 3.69\% versus 4.31\%, compared with 2.63\% versus 2.80\% without events. The variant uses the same model and inputs to forecast the entire horizon, so this comparison evaluates the full adaptive workflow.

\textbf{Interaction recovery.} To complement forecasting results under real-event co-occurrence, this controlled experiment tests whether forecasts recover specified interaction effects. We construct synthetic event cases with four conditions for each GEFCom2014 base case: $0$ retains the original future targets without adding events; $A$ and $B$ each add a different relevant event; and $AB$ includes both. We modify future targets according to the events and supply the corresponding event descriptions.

An event's effect is its load change relative to condition $0$. Interaction is the change under $AB$ minus the sum of the changes under $A$ and $B$. Zero interaction means that the combined effect equals this sum; nonzero interaction means a larger or smaller combined effect. WAPE measures load forecast accuracy under each condition. Interaction estimation error (IEE) compares the interaction computed from the four forecasts, $\widehat{\Delta}_{AB,k}$, with its target counterpart:
\begin{equation}
\begin{aligned}
\Delta_{AB,k} &= y_{AB,k}-y_{A,k}-y_{B,k}+y_{0,k},\\
\mathrm{IEE} &= \frac{\sum_k |\widehat{\Delta}_{AB,k}-\Delta_{AB,k}|}
{\sum_k |y_{0,k}|},
\end{aligned}
\label{eq:iee-main}
\end{equation}
where $y_{s,k}$ is the target load under condition $s$ at step $k$. Both sums cover the forecast horizon; IEE is normalized by the sum of absolute loads under condition $0$ and reported as a percentage.

Figure~\ref{fig:multi-event-interaction-demo}(a) reports WAPE under each condition; under $AB$, ElecLoad achieves 3.20\% versus 3.94\% for the direct-only variant. Figure~\ref{fig:multi-event-interaction-demo}(b) groups cases by zero or nonzero target interaction; in the nonzero group, ElecLoad's IEE is 0.57\% versus 1.12\% for this variant. In this controlled comparison, ElecLoad is more accurate in both load forecasting and interaction estimation.

\textbf{Generalization.} Figure~\ref{fig:multi-event-generalization-demo}(c,d) reports WAPE (c) and IEE (d) under Seen (combinations encountered during training), New overlap (seen combinations with unseen temporal overlap), and New pair (unseen combinations). ElecLoad achieves the lowest values of both metrics among the four methods in every setting. Appendix~\ref{app:synthetic-events} provides construction details.

Appendix~\ref{app:irrelevant-text} provides a supplementary analysis of how adding irrelevant event text affects load forecasts.

\begin{table}[!t]
 \caption{Ablations of routing, refinement, and reconciliation.}
 \label{tab:planning-utility-demo}
 \centering\fontsize{6.5}{7.8}\selectfont
 \setlength{\tabcolsep}{1pt}
 \renewcommand{\arraystretch}{1.12}
 \begin{tabular*}{\columnwidth}{@{\extracolsep{\fill}}ll*{9}{c}@{}}
 \toprule
 \multirow{2}{*}{Group} & \multirow{2}{*}{Variant} & \multicolumn{3}{c}{GEFCom2014} & \multicolumn{3}{c}{PJM} & \multicolumn{3}{c}{LCL} \\
 \cmidrule(lr){3-5}\cmidrule(lr){6-8}\cmidrule(lr){9-11}
 & & $E$ & $C$ & $G$ & $E$ & $C$ & $G$ & $E$ & $C$ & $G$ \\
 \midrule
\multirow{3}{*}{Routing} & w/ direct only & 7.82 & 7.8 & --- & 3.48 & 7.0 & --- & 42.91 & 6.5 & --- \\
 & w/ segmented only & 6.97 & 24.0 & 0.052 & 3.39 & 22.0 & 0.006 & 45.36 & 20.0 & -0.181 \\
 & ElecLoad (Full) & \textbf{6.68} & 18.7 & \textbf{0.105} & \textbf{3.13} & 15.3 & \textbf{0.042} & \textbf{41.94} & 12.7 & \textbf{0.156} \\
\midrule
\multirow{3}{*}{Refinement} & w/o refinement & 7.24 & 15.0 & --- & 3.37 & 12.2 & --- & 44.63 & 10.0 & --- \\
 & w/ random refinement & 6.94 & 18.6 & 0.083 & 3.32 & 15.2 & 0.017 & 43.06 & 12.8 & 0.561 \\
 & ElecLoad (Full) & \textbf{6.68} & 18.7 & \textbf{0.151} & \textbf{3.13} & 15.3 & \textbf{0.077} & \textbf{41.94} & 12.7 & \textbf{0.996} \\
\midrule
\multirow{3}{*}{Reconciliation} & w/o reconciliation & 6.77 & 17.6 & --- & 3.15 & 14.4 & --- & 42.08 & 12.1 & --- \\
 & w/ forced reconciliation & 6.73 & 19.5 & 0.021 & 3.19 & 16.2 & -0.022 & 42.31 & 13.7 & -0.144 \\
 & ElecLoad (Full) & \textbf{6.68} & 18.7 & \textbf{0.082} & \textbf{3.13} & 15.3 & \textbf{0.022} & \textbf{41.94} & 12.7 & \textbf{0.233} \\
\bottomrule
 \end{tabular*}
 \par\vspace{2pt}
\end{table}

\subsection{Planning and Refinement Analysis}
\label{sec:planning-refinement}

\textbf{Accuracy--cost trade-off.} Table~\ref{tab:planning-utility-demo} evaluates how routing, refinement, and reconciliation affect forecasting accuracy and token cost. $E$ is WAPE (\%), and $C$ is the mean total number of input and output tokens per case, in thousands. We define $G=(E_{\mathrm{ref}}-E)/(C-C_{\mathrm{ref}})$ using each group's first row as the reference for each dataset. It measures the WAPE reduction in percentage points per additional 1,000 tokens; ``---'' denotes an undefined value.

The direct-only variant always uses direct forecasting. The segmented\hyp{}only variant always partitions the horizon into segments. It retains event-guided boundaries, selective refinement, and boundary reconciliation. Figure~\ref{fig:planning-accuracy-cost-demo} plots WAPE against normalized token cost on all GEFCom2014 test cases. For each case, we divide its token count by that of its direct run, then average over cases. Panel~(a) compares ElecLoad with the direct-only and segmented-only variants. Panel~(b) compares refinement and reconciliation with every case on the segmented path. The refinement and reconciliation groups in Table~\ref{tab:planning-utility-demo} retain adaptive routing.

Adaptive routing achieves lower WAPE with fewer tokens than the segmented-only variant on all three datasets. On GEFCom2014, ElecLoad reduces WAPE from 6.97\% to 6.68\% while using 22.1\% fewer tokens. Relative to direct-only forecasting, it reduces WAPE from 7.82\% to 6.68\% at a higher cost of 18.7k versus 7.8k tokens. Selective refinement also achieves lower WAPE than random refinement at similar token cost across all three datasets; on GEFCom2014, the comparison is 6.68\% versus 6.94\% at 18.7k versus 18.6k tokens. Boundary reconciliation achieves lower error with fewer tokens than forced reconciliation. Together, these comparisons support adapting both the forecasting path and the allocation of refinement calls within the evaluated workflow.

\textbf{Case study.} Figure~\ref{fig:planning-case-demo}(a) shows overlapping heatwave and demand response events. Event analysis guides T1 to partition the horizon into S1--S3; S2 covers their overlap. T2 independently forecasts each segment, forming ``T2 Initial'', which overestimates load in S2. In panel~(b), T3 provides feedback; T2 revises S2 values, and the main agent reassesses the full forecast to obtain ``After T3''. T4's boundary reconciliation locally adjusts forecast values near segment boundaries, producing ``Final'' and complementing refinement focused on the overestimated segment.

\section{Conclusion}

We present ElecLoad, which uses multi-event analysis to guide adaptive electricity load forecasting so that the forecasting process follows changes in event effects over time. Experiments on day-ahead and multi-day forecasting show that ElecLoad performs best in most metric comparisons; its error reduction over the direct-only variant, which uses the same model and inputs, is larger under multi-event conditions. Ablation analysis shows that the relative performance of direct and segmented forecasting varies across datasets, whereas the adaptive forecasting process achieves lower average error than either fixed forecasting path. Selective refinement yields lower error than random refinement at similar token cost across all three evaluated datasets. These results indicate that event analysis can effectively inform which periods to forecast together and which segments require refinement.

\bibliographystyle{ACM-Reference-Format}
\bibliography{main}
\clearpage
\appendix
\setlength{\emergencystretch}{2em}
\raggedbottom
\section{Evaluation Construction}
\label{app:controlled}\label{sec:controlled}\label{app:recorded-comparators}

\subsection{Data Sources}
\label{app:data-sources}
GEFCom2014 load-track series~\cite{hong2016probabilistic} provide original cases without added events and the basis for our synthetic event cases.
Nord combines national load from ENTSO-E Power Statistics~\cite{entsoePowerStats} with \href{https://umm.nordpoolgroup.com/}{Nord Pool Urgent Market Messages}.
PJM uses metered hourly native load from PJM Data Miner~2~\cite{pjmDataMiner}.
PJM and SPP (SWPP) use EIA-930 balancing-area demand~\cite{eiaGridMonitor} obtained through \href{https://docs.catalyst.coop/pudl/en/stable/data_sources/eia930.html}{Catalyst Cooperative PUDL}.
Both use National Weather Service (NWS) weather alerts from the \href{https://mtarchive.geol.iastate.edu/}{Iowa Environmental Mesonet archive}; PJM also uses \href{https://emergencyprocedures.pjm.com/ep/pages/dashboard.jsf}{PJM Emergency Procedures} alerts.

LCL uses UK Power Networks smart-meter consumption and tariffs from the London Datastore~\cite{lclSmartMeter}, with daily tariff notices as events.
iFlex provides household load, experiment days, and research price schedules~\cite{iflexDataset}.
AEMO uses NEMWeb operational demand~\cite{aemoNemweb} and \href{https://www.nemweb.com.au/REPORTS/CURRENT/Market_Notice/}{market notices}.
ERCOT uses its Hourly Load Data Archives~\cite{ercotHourlyLoad}, \href{https://www.ercot.com/news/releases}{news releases}, operating condition notices, \mbox{and NWS weather alerts.}

\subsection{Temporal Splits and Window Selection}
\label{app:splits-windows}
ID cases are assigned chronologically by the date of the first predicted hour. Table~\ref{tab:app-date-ranges} summarizes these dates across retained cases, combining both horizons where used. The ranges describe selected windows without implying continuous coverage. Dates use UTC except for GEFCom2014, which retains its source time axis; ERCOT includes all 85 cases in Table~\ref{tab:data}. Nord uses origins before 2023 for training, 2023 for validation, and 2024 for testing; PJM and SPP use origins before 2023, in 2023--2024, and from 2025 onward, respectively. LCL and iFlex split the eligible schedule dates chronologically, keeping all household views of a date in the same split. Their validation and test households also appear in training, so these evaluations concern forecasting later dates for known households.

Synthetic variants inherit the split of their original load window. No GEFCom2014 parent window is shared across training, validation, and testing. Nord, PJM, and SPP additionally keep explicitly linked notices and revisions in the same split. For distinct original windows, selected ID forecast targets do not overlap on the same physical load series, including across splits. For later origins, input histories may reuse earlier cases' target hours \mbox{once these are observed.}

\begin{table}[t]
\caption{Calendar date ranges by data source and split.}
\label{tab:app-date-ranges}
\centering\small
\setlength{\tabcolsep}{3pt}
\renewcommand{\arraystretch}{1.15}
\begin{tabular*}{\columnwidth}{@{\extracolsep{\fill}}lll@{}}
\toprule
Source & Split & Date range \\
\midrule
\multirow{3}{*}{GEFCom2014} & Train & 2005-01-05--2009-10-31 \\
 & Val. & 2009-11-08--2010-11-11 \\
 & Test & 2010-11-19--2011-11-26 \\
\addlinespace[2pt]
\multirow{3}{*}{Nord} & Train & 2015-06-30--2022-12-30 \\
 & Val. & 2023-01-01--2023-12-05 \\
 & Test & 2024-01-01--2024-11-19 \\
\addlinespace[2pt]
\multirow{3}{*}{PJM} & Train & 2004-05-18--2022-12-10 \\
 & Val. & 2023-01-01--2024-11-06 \\
 & Test & 2025-01-01--2026-07-18 \\
\addlinespace[2pt]
\multirow{3}{*}{SPP} & Train & 2015-07-08--2022-12-26 \\
 & Val. & 2023-01-01--2024-06-22 \\
 & Test & 2025-01-01--2026-07-14 \\
\addlinespace[2pt]
\multirow{3}{*}{LCL} & Train & 2013-01-04--2013-11-20 \\
 & Val. & 2013-11-21--2013-12-09 \\
 & Test & 2013-12-10--2013-12-26 \\
\addlinespace[2pt]
\multirow{3}{*}{iFlex} & Train & 2020-02-11--2021-02-12 \\
 & Val. & 2021-02-15--2021-02-24 \\
 & Test & 2021-03-05--2021-03-25 \\
\midrule
AEMO (OOD) & Test & 2025-08-08--2026-09-24 \\
ERCOT (OOD) & Test & 2021-02-12--2026-08-26 \\
\bottomrule
\end{tabular*}
\end{table}

\paragraph{Window selection.}
All cases use hourly load with the history and forecast lengths in Table~\ref{tab:data}. GEFCom2014 windows are drawn from a 96-hour origin grid. Nord, PJM, and SPP select windows for coverage of source entities, calendar quarters, and event-description counts; LCL and iFlex balance household and schedule-date coverage. The retained origins therefore need not have uniform spacing. AEMO and ERCOT are reserved for OOD evaluation, with aggregation defined below.

\subsection{Forecast-Origin Information}
\label{app:forecast-inputs}
Inputs contain observed load history, available covariates, and event announcements available at the forecast origin. Forecast-horizon temperatures in GEFCom2014 are weather forecasts available at that origin. LCL tariff notices and iFlex research price schedules provide next-day planned conditions. An event's effective interval specifies its alignment with the load horizon, separately from the announcement's availability time. Label-assisted teacher demonstrations use training cases; held-out load labels are used for evaluation.

\subsection{Synthetic Event Construction}
\label{app:synthetic-events}
Programmes A--F are industrial pause $(-0.12,0)$, demand response $(-0.20,0.5)$, commercial closure $(-0.10,0)$, commissioning $(0.12,0)$, deferred demand $(-0.16,1)$ and restart $(0.10,0)$, respectively; pairs specify $(b,\rho)$. Table~\ref{tab:app-events} counts forecast-window effects. With duration $d\in\{6,12\}$, intensity $q\in\{0.75,1,1.25\}$ and start $s$,
\begin{equation}
\delta_t=\begin{cases}bqz_t,&s\le t<s+d,\\
-\rho bq\sum_{u=s}^{s+d-1}z_u/d,&s+d+2\le t<s+2d+2,\\
0,&\text{otherwise}.\end{cases}
\end{equation}
where $b$ is the signed fraction, $\rho$ the return fraction, and $z$ the curve the programme acts on.
Independent composition sums parent-curve responses; shared composition applies events successively by start time, breaking ties alphabetically. History-changing families are evaluated separately. Eight matched conditions define
\begin{align}
\Delta_{ABC}&=\sum_{S\subseteq\{A,B,C\}}(-1)^{3-|S|}y_S,\\
\mathrm{IEE}_3&=100\frac{\|\widehat\Delta_{ABC}-\Delta_{ABC}\|_1}{\|y_0\|_1}.
\end{align}
where $y_{\varnothing}=y_0$, hats denote predictions, and IEE$_3$ is a percentage.

GEFCom2014 counts each programme once over the union of its direct and recovery impact windows; overlap percentages in Table~\ref{tab:app-events} use cases with multiple relevant events.

\begin{table}[!htbp]
\caption{GEFCom2014 case composition ($L=96$, $H=96$).}
\label{tab:app-events}
\centering\fontsize{6.5}{7.8}\selectfont
\setlength{\tabcolsep}{1pt}
\renewcommand{\arraystretch}{1.12}
\begin{tabular*}{\columnwidth}{@{\extracolsep{\fill}}lrrrrrrl@{}}
\toprule
Split & $N$ & 0 & H & I & 1 & $\geq$2 & Overlap \\
\midrule
Train & 640 & 105 & 15 & 80 & 250 & 190 & 130 (68.4) \\
Val. & 100 & 15 & 2 & 20 & 36 & 27 & 20 (74.1) \\
ID test & 200 & 30 & 4 & 40 & 72 & 54 & 40 (74.1) \\
\bottomrule
\end{tabular*}
\par\vspace{3pt}
{\raggedright 0: original; H: history-only programmes; I: irrelevant-only; 1/$\geq$2: one/multiple future-relevant programmes. Overlap: count (\%).\par}
\end{table}

\subsection{Event Statistics and Evaluation Groups}
\label{app:event-groups}
Table~\ref{tab:app-event-records} counts input event descriptions, with each announcement or daily price schedule counted once. Two event descriptions overlap when neither is explicitly marked irrelevant and their specified effective intervals intersect for a positive duration within \mbox{the forecast window.}

\begin{table}[!htbp]
\caption{Test-set event descriptions and temporal overlap.}
\label{tab:app-event-records}
\centering\fontsize{6.5}{7.8}\selectfont
\setlength{\tabcolsep}{0.5pt}
\renewcommand{\arraystretch}{1.12}
\begin{tabular*}{\columnwidth}{@{\extracolsep{\fill}}llrrrrrlr@{}}
\toprule
& & & \multicolumn{4}{c}{Event descriptions} & & \\
\cmidrule(lr){4-7}
Source & $L/H$ & $N$ & 0 & 1 & 2 & $\geq$3 & Overlap & Incomp. \\
\midrule
\multicolumn{9}{l}{ID test} \\
GEFCom2014 original & 96/96 & 30 & 30 & 0 & 0 & 0 & 0 & 0 \\
GEFCom2014 synthetic & 96/96 & 170 & 0 & 102 & 58 & 10 & 40 (23.5\%) & 0 \\
Nord & 168/24 & 96 & 0 & 54 & 30 & 12 & 30 (31.2\%) & 16 \\
Nord & 96/96 & 24 & 0 & 15 & 5 & 4 & 3 (12.5\%) & 11 \\
PJM & 168/24 & 112 & 35 & 40 & 14 & 23 & 21 (18.8\%) & 37 \\
PJM & 96/96 & 28 & 8 & 10 & 2 & 8 & 7 (25.0\%) & 6 \\
SPP & 168/24 & 48 & 22 & 14 & 5 & 7 & 12 (25.0\%) & 0 \\
SPP & 96/96 & 12 & 5 & 3 & 1 & 3 & 4 (33.3\%) & 0 \\
LCL & 168/24 & 48 & 12 & 24 & 9 & 3 & 5 (10.4\%) & 0 \\
iFlex & 168/24 & 32 & 0 & 32 & 0 & 0 & 0 & 0 \\
\midrule
\multicolumn{9}{l}{OOD test} \\
AEMO & 96/96 & 150 & 0 & 0 & 34 & 116 & 42 (28.0\%) & 120 \\
ERCOT & 168/24 & 85 & 0 & 20 & 5 & 60 & 63 (74.1\%) & 0 \\
\bottomrule
\end{tabular*}
\par\vspace{3pt}
{\raggedright $N$: test cases in the row. $0/1/2/{\geq}3$: case counts by number of event descriptions. Overlap percentages use all $N$ cases. Incomp.: cases containing at least one event description with an incomplete interval. ERCOT description counts cover all 85 cases; the main performance evaluation uses the 71 system-level cases.\par}
\end{table}

\paragraph{Event-count groups.}
Table~\ref{tab:app-pjm-overlap} groups cases by the number of distinct events whose windows intersect the forecast horizon for a positive duration. PJM and SPP use recorded effective intervals, counting repeated notices of one event once. Both show larger WAPE reductions over direct-only forecasting with multiple events.

\input{tables/pjm_event_overlap}

\paragraph{Event-relevance groups.}
Case counts for Table~\ref{tab:event-strata-demo} are as follows. GEFCom2014: 30 No Event, 72 Single Event, 54 Multiple Events, 40 Irrelevant Events; four history-only cases fall outside these conditions. PJM: 35 No Event, 33 Single Event, 28 Multiple Events, and 16 Irrelevant Events. LCL: 12 cases per condition.

\subsection{Evaluation Aggregation}
\label{app:aggregation}
Reported metrics use the forecast actually delivered for every case, including the recorded fallback after failure. Dataset results average case errors; for AEMO, regional errors are first averaged within each forecast origin, then across origins. The 85 ERCOT cases comprise 71 system-level cases for the main performance evaluation and 14 regional sensitivity cases. The latter are excluded from the main performance metrics. WAPE and NRMSE retain near-zero targets without denominator offsets; zero denominators are undefined. LCL's low-consumption windows can yield large percentage errors with equal weight. We checked the reported results for consistency with the experimental code, configurations, and data.

\section{Training and Runtime Definitions}
\label{app:runtime}

\subsection{Reasoning Training}
\label{app:reasoning-training}
The teacher model is DeepSeek-v4-flash~\cite{deepseek2026v4flash}. Reasoning RL uses group relative policy optimization (GRPO)~\cite{shao2024deepseekmath} with the reward below and selects its checkpoint by mean validation reward.

Reasoning uses clipped WAPE $W$ and NRMSE $N$. Let $b_*$ be the smaller available historical or HGB reference loss, and $[x]_0^1=\min(1,\max(0,x))$:
\begin{align}
L&=0.9\min(W,4)+0.1\min(N,4),\\
d_*&=b_*-L,\\
g&=d_*/\max(b_*,0.1),\\
B&=0.1\min\!\left(\left[\frac{g-0.05}{0.20}\right]_0^1,\left[\frac{d_*-0.02}{0.08}\right]_0^1\right),\\
s&=\frac{\operatorname{mean}|\hat y|}{\max(\operatorname{mean}|y|,\operatorname{sd}(y),1)+10^{-8}},\\
x&=\frac{\max|\hat y|}{\max(\max|y|,\operatorname{sd}(y),1)+10^{-8}},\\
P_s&=-0.2\,\mathbf1_{s>10}\min(1,\log_{10}(s)/3),\\
P_x&=-0.8\,\mathbf1_{x>100}\min(1,\log_{10}(x/100+1)),\\
R_{\rm reasoning}&=\begin{cases}-L+B+P_s+P_x,&\text{valid output},\\-6,&\text{otherwise}.\end{cases}
\end{align}
The bonus is zero without a reference. Historical candidates repeat the last value, recent horizon mean, or an autocorrelation-qualified seasonal pattern (threshold 0.5).

\subsection{Agent SFT}
\label{app:agent-sft}
Agent SFT minimizes Eq.~\ref{eq:agent_sft}, supervising each call's current response in its own context.

For a batch $\mathcal D$ of demonstration episodes, SFT averages valid response-token losses within episodes and then across episodes:
\begin{equation}
\mathcal{L}_{\mathrm{SFT}}(\theta)
=-\frac{1}{|\mathcal D|}\sum_{e\in\mathcal D}
\frac{1}{|\mathcal T_e|}\sum_{t\in\mathcal T_e}
\log\pi_\theta(a_{e,t}\mid h_{e,t}),
\label{eq:agent_sft}
\end{equation}
where $\pi_\theta$ is the shared model's conditional token probability, $\mathcal T_e$ indexes all valid supervised response tokens in $e$, and $a_{e,t}$ is the target token. Each $h_{e,t}$ contains the corresponding call's input and response prefix before $t$.

\subsection{Agent RL}
\label{app:agent-rl}
Agent RL uses clipped policy ratios~\cite{schulman2017proximal} and KL regularization toward frozen agent SFT weights. Its checkpoint maximizes the mean selected-path validation score across all attempted cases:
\begin{equation}
R(e)=U(e)-F(e),
\label{eq:validation-score}
\end{equation}
where $U$ follows Eq.~\ref{eq:reward}, and $F(e)$ equals one for failure to complete the selected path normally or for fallback delivery, and zero otherwise. Normal completion requires valid outputs from every required stage after any allowed repairs. Scores use the forecasts actually delivered, including fallbacks, and all incurred call costs, \mbox{including failed attempts.}

\paragraph{Role-specific credit.}
Training uses two independently sampled T1 executions per case, with paired direct and segmented evaluations. T1 compares the utility of its selected path with the other sample's selected path and with its own alternative path. Initial T2 forecasts retain the same-path utility comparison between samples and add improvement over a last-day repetition baseline. T2 refinement and T4 receive credit for immediate error reduction minus call cost. T3 receives credit for error reduction through the subsequent refinement loop, minus its own and subsequent call costs:
\begin{align}
A_{\rm T1}&=U_{\rm selected}-U^{\rm sibling}_{\rm selected}\nonumber\\
&\quad+\tfrac12 I_{\rm pair}(U_{\rm selected}-U_{\rm other}),\\
A_{\rm init,local}&=10\frac{\sum_{t\in S}(|y_t-y^{\rm rep}_t|-|y_t-\hat y_t|)}{\|y\|_1},\\
A_{\rm refine/T4}&=10 I_{\rm accepted}(W_{\rm before}-W_{\rm after})\nonumber\\
&\quad-0.01C_{\rm attempt}/C_{\rm ref},\\
A_{\rm T3}&=10 I_{\rm accepted}(W_{\rm before}-W_{\rm preT4})\nonumber\\
&\quad-0.01(C_{\rm own}+C_{\rm suffix})/C_{\rm ref}.
\end{align}
Here, $U$ is the utility in Eq.~\ref{eq:reward}; $U_{\rm selected}$ and $U_{\rm other}$ refer to the selected and alternative paths, and $U^{\rm sibling}_{\rm selected}$ to the other independent sample's selected path. $I_{\rm pair}=1$ when the T1 attempt is accepted and both paths complete normally. For initial T2 forecasts, $S$ is the assigned segment (the full horizon on the direct path), $y^{\rm rep}$ repeats the last 24 observed loads, and $\|y\|_1$ sums absolute target loads over the full horizon.

$W$ denotes full-horizon WAPE as a fraction. Subscripts before and after denote predictions immediately before and after an update; preT4 denotes the last retained prediction after the refinement loop and before T4. $I_{\rm accepted}=1$ when the attempt passes output validation. $C_{\rm attempt}$ and $C_{\rm own}$ count input and output tokens of the current refinement/T4 and T3 attempts, respectively; $C_{\rm suffix}$ counts subsequent T2/T3 attempts before T4. The frozen per-case reference $C_{\rm ref}$ uses Agent SFT with a direct plan and one full-horizon prediction, or an identified same-case Reasoning fallback.

After two refinements, T3 must stop and receives no task credit. An invalid T3 attempt incurs its own cost but receives no quality gain or subsequent-call cost. If a call remains invalid after its repair, both attempts receive a $-1$ output-validity signal, optimized with a separate clipped policy objective.

\subsection{Runtime Settings and Output Handling}
\label{app:runtime-settings}

\begin{table}[!htbp]
\caption{Runtime settings.}
\label{tab:runtime-settings}
\centering\small
\begin{tabular}{@{}p{0.60\columnwidth}p{0.34\columnwidth}@{}}
\toprule
Setting & Value \\
\midrule
Agent RL batch; epochs; learning rate & $16;\ 2;\ 10^{-6}$ \\
Train/validation; PPO clip; KL & $1800/300;\ 0.2;\ 0.001$ \\
Prompt/response/context tokens & $32768/4096/36864$ \\
Train/validation temperature & $1/0$ \\
Max segments; minimum steps & $4;\ 6$ \\
Refinements; repairs/call & $2;\ 1$ \\
Edit steps per patch; total & $4;\ 8$ \\
\bottomrule
\end{tabular}
\end{table}

\paragraph{Output validation and fallback.}
ElecLoad checks response structure, numerical finiteness, and forecast length, allowing one repair per invalid call. If execution still fails, it returns the latest complete forecast when available; otherwise, it repeats the last 24 observed loads to cover the horizon. Evaluation uses the delivered forecast, and token costs include failed calls and repairs.

\subsection{Baseline Implementations}
\label{app:baseline-implementations}
The baselines use the same evaluation cases through their supported input interfaces. TimeReasoner uses DeepSeek-v4-flash. CastFSR uses a training-free API coordinator; MemCast uses a training-free variant without visual analysis. The EventTSF adaptation normalizes load history and uses event text. ITA-LF retains its trained checkpoint and constrains decoding to the required output format and horizon length.

\section{Additional Analyses}
\label{app:additional-analyses}

Unless stated otherwise, tables use $L=96,H=96$ for GEFCom2014 and $L=168,H=24$ for PJM/LCL. kTok denotes thousands of tokens per case; component and execution ablations \mbox{measure cost on GEFCom2014.}

\subsection{Model Scale and Training Ablations}
\label{app:model-training-ablations}
The variant without Agent training in Table~\ref{tab:app-training} retains ElecLoad's execution protocol. In Table~\ref{tab:app-training}, 4B/9B WAPEs differ by at most \mbox{0.13 percentage points.}

\begin{table}[!htbp]
\caption{Model scale and training.}
\label{tab:app-training}
\centering\fontsize{6.5}{7.8}\selectfont
\setlength{\tabcolsep}{1pt}
\renewcommand{\arraystretch}{1.12}
\begin{tabular*}{\columnwidth}{@{\extracolsep{\fill}}lrrr@{}}
\toprule
Setting & GEFCom2014 & PJM & LCL \\
\midrule
ElecLoad 0.8B & 8.14 & 3.37 & 47.36 \\
ElecLoad 4B & 6.68 & 3.13 & 41.94 \\
ElecLoad 9B & 6.63 & 3.16 & 41.81 \\
w/o Reasoning init. & 7.42 & 3.46 & 45.08 \\
w/o Agent training & 8.05 & 3.83 & 47.18 \\
w/o Agent SFT & 7.19 & 3.42 & 44.28 \\
w/o Agent RL & 7.06 & 3.33 & 43.35 \\
Reasoning: SFT+RL & 7.37 & 3.42 & 44.32 \\
Reasoning w/o SFT & 8.16 & 3.87 & 48.18 \\
Reasoning w/o RL & 7.68 & 3.63 & 45.87 \\
\bottomrule
\end{tabular*}
\end{table}

\subsection{Training-Component Ablations}
\label{app:component-ablations}
Table~\ref{tab:app-objectives} uses the Reasoning SFT+RL and ElecLoad 4B rows of Table~\ref{tab:app-training} as references (ElecLoad: 18.7k tokens per GEFCom2014 case). Removing the numerical term drops $-L$ while $B$ still uses $L$; removing the format penalty assigns invalid outputs reward 0. Local quality credit comprises quality terms for initial T2 forecasting, T2 refinement, T3, and T4. On GEFCom2014, cost credit saves 8.9k tokens per case for 0.08 points higher WAPE, as shown in Table~\ref{tab:app-objectives}.

\begin{table}[!htbp]
\caption{Training-component ablations.}
\label{tab:app-objectives}
\centering\fontsize{6.5}{7.8}\selectfont
\setlength{\tabcolsep}{1pt}
\renewcommand{\arraystretch}{1.12}
\begin{tabular*}{\columnwidth}{@{\extracolsep{\fill}}ll*{4}{c}@{}}
\toprule
Group & Removed component & GEFCom2014 & PJM & LCL & kTok \\
\midrule
\multirow{5}{*}{Reasoning} & Numerical term & 8.13 & 3.83 & 48.01 & -- \\
 & Improvement bonus & 7.52 & 3.53 & 44.97 & -- \\
 & Scale penalty & 7.62 & 3.66 & 47.27 & -- \\
 & Extreme penalty & 7.56 & 3.52 & 46.87 & -- \\
 & Format penalty & 7.45 & 3.52 & 45.03 & -- \\
\midrule
\multirow{5}{*}{Agent} & Final accuracy credit & 7.48 & 3.58 & 45.72 & 17.7 \\
 & Cost credit & 6.60 & 3.12 & 41.87 & 27.6 \\
 & Contract credit & 7.05 & 3.36 & 44.12 & 21.39 \\
 & Local quality credit & 7.15 & 3.43 & 44.65 & 19.11 \\
 & Paired route credit & 7.07 & 3.39 & 43.93 & 22.21 \\
\bottomrule
\end{tabular*}
\end{table}

\subsection{Execution Variants}
\label{app:execution-variants}
In Table~\ref{tab:app-execution}, the default allows four segments, two refinements and eight edited steps. The four forced segments retain event-guided boundaries for $H=96$ and contain six steps each for $H=24$; forced refinements disable stopping until the specified count. First-segment refinement always selects $S_1$ under the same budget. Regeneration repeats all segments from original inputs and existing reports without feedback. On GEFCom2014, the default workflow has lower WAPE and cost than refining all segments once or regenerating them. Four refinements slightly lower WAPE \mbox{at higher cost.}

\begin{table}[!htbp]
\caption{Execution variants.}
\label{tab:app-execution}
\centering\fontsize{6.5}{7.8}\selectfont
\setlength{\tabcolsep}{1pt}
\renewcommand{\arraystretch}{1.12}
\begin{tabular*}{\columnwidth}{@{\extracolsep{\fill}}lrrrr@{}}
\toprule
Setting & GEFCom2014 & PJM & LCL & kTok \\
\midrule
Default & 6.68 & 3.13 & 41.94 & 18.7 \\
At most 2 segments & 6.95 & 3.26 & 42.05 & 15.9 \\
At most 3 segments & 6.73 & 3.14 & 41.98 & 17.5 \\
Force 4 segments & 7.06 & 3.33 & 44.93 & 26.1 \\
Refine first segment & 6.98 & 3.35 & 43.46 & 18.5 \\
Force 1 refinement & 6.85 & 3.21 & 42.73 & 17.3 \\
Force 2 refinements & 6.73 & 3.22 & 42.24 & 21.4 \\
Allow 4 refinements & 6.63 & 3.15 & 41.93 & 22.8 \\
Refine all segments once & 6.73 & 3.18 & 42.08 & 26.1 \\
Regenerate all segments & 6.84 & 3.29 & 42.83 & 29.5 \\
Edit limit: 4 steps & 6.73 & 3.17 & 42.06 & 18.2 \\
Edit limit: 12 steps & 6.72 & 3.18 & 42.05 & 19.5 \\
\bottomrule
\end{tabular*}
\end{table}

\subsection{Irrelevant-Text Effects}
\label{app:irrelevant-text}
\begin{figure}[!htbp]
 \centering
 \includegraphics[width=\columnwidth]{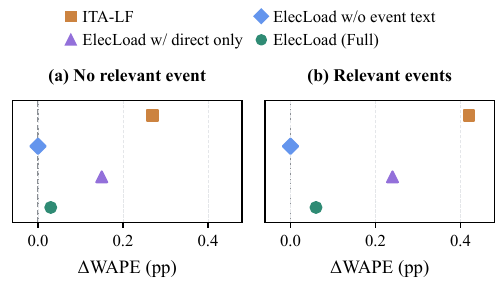}
 \caption{Mean WAPE change after adding irrelevant text on GEFCom2014 (pp: percentage points).}
 \label{fig:multi-event-distractors-demo}
\end{figure}

We measure how irrelevant event text changes forecasting error by adding it to GEFCom2014 cases while keeping the load history, covariates, target load, and any existing relevant event descriptions fixed. For each method, we report its mean WAPE after the addition minus its mean WAPE before, in percentage points, on the same cases; positive values mean higher error. Figure~\ref{fig:multi-event-distractors-demo}(a) adds the text to cases without relevant events; Figure~\ref{fig:multi-event-distractors-demo}(b) adds it alongside existing relevant event descriptions. Among the methods that use event text, ElecLoad has the smallest increase in both settings.

The irrelevant-text analysis in Table~\ref{tab:app-controlled} uses a separate set of 48 cases per cell.

\begin{table}[!htbp]
\caption{Irrelevant-text effects ($L=96$, $H=96$).}
\label{tab:app-controlled}
\centering\fontsize{6.5}{7.8}\selectfont
\setlength{\tabcolsep}{1pt}
\renewcommand{\arraystretch}{1.12}
\begin{tabular*}{\columnwidth}{@{\extracolsep{\fill}}lrrrr@{}}
\toprule
& \multicolumn{2}{c}{No events} & \multicolumn{2}{c}{Relevant events} \\
Type & E & D & E & D \\
\midrule
Administrative & +0.012 & -0.023 & +0.045 & -0.015 \\
Wrong target & +0.032 & -0.013 & +0.061 & -0.011 \\
Outside horizon & -0.005 & +0.027 & -0.005 & -0.007 \\
Cancelled & +0.028 & +0.057 & +0.045 & -0.086 \\
Superseded & -0.032 & -0.033 & -0.043 & -0.043 \\
\bottomrule
\end{tabular*}
\par\vspace{3pt}
{\raggedright E: ElecLoad; D: direct-only.\\
Values: paired after-minus-before WAPE changes (percentage~points).\par}
\end{table}

\subsection{Paired WAPE Differences}
\label{app:paired-differences}
Table~\ref{tab:app-diagnostics} reports paired WAPE differences between ElecLoad and direct-only forecasting; negative values favor ElecLoad. For GEFCom2014 and PJM, the 95\% intervals are below zero; the LCL interval includes zero.

\begin{table}[!htbp]
\caption{Paired WAPE differences.}
\label{tab:app-diagnostics}
\centering\fontsize{6.5}{7.8}\selectfont
\setlength{\tabcolsep}{1pt}
\renewcommand{\arraystretch}{1.12}
\begin{tabular*}{\columnwidth}{@{\extracolsep{\fill}}lrrrr@{}}
\toprule
Source & $\Delta$WAPE & 95\% interval & kTok & P99 \\
\midrule
GEFCom2014 & -1.14 & [-1.31, -0.98] & 18.7 & 16.18 \\
PJM & -0.35 & [-0.58, -0.12] & 15.3 & 8.50 \\
LCL & -0.97 & [-1.95, +0.07] & 12.7 & 79.88 \\
\bottomrule
\end{tabular*}
\par\vspace{3pt}
{\raggedright $\Delta$WAPE: ElecLoad minus direct-only (percentage points). \\
P99: 99th percentile of ElecLoad case~WAPE~(\%).\par}
\end{table}

\clearpage

\end{document}

%% file: algorithms/workflow.tex
\begin{algorithm}[t]
\caption{ElecLoad inference workflow}
\label{alg:elecload-workflow}
\small
\DontPrintSemicolon
\SetAlgoVlined
\SetKw{Break}{break}
\SetKwFunction{Tone}{T1}
\SetKwFunction{Ttwo}{T2}
\SetKwFunction{Tthree}{T3}
\SetKwFunction{Tfour}{T4}
\SetKwFunction{Concat}{Concat}
\SetKwFunction{ApplyPatch}{ApplyPatch}
\KwIn{$\mathbf x=(\mathbf y_{-L:-1},\mathcal C)$, horizon $H$, refinement budget $R_{\max}$}
\KwOut{Complete load forecast $\hat{\mathbf y}\in\mathbb R^H$}
\tcp{T1: event-guided planning}
$(\mathcal A,m,\mathcal S)\gets$ \Tone{$\mathbf x,H$}\;
\If{$m=\texttt{DIRECT}$}{
  \KwRet{\Ttwo{$\mathbf x,\mathcal A,[0,H)$}}\;
}
\tcp{T2: independent forecasting}
\ForEach{$S_j\in\mathcal S$}{
  $\hat{\mathbf y}_{S_j}\gets$ \Ttwo{$\mathbf x,\mathcal A,S_j$}\;
}
$\hat{\mathbf y}\gets$ \Concat{$\hat{\mathbf y}_{S_1},\ldots,\hat{\mathbf y}_{S_K}$}\;
\tcp{T3: selective refinement}
\For{$r=0,\ldots,R_{\max}$}{
  $(a,j,f)\gets$ \Tthree{$\mathbf x,\mathcal A,\mathcal S,\hat{\mathbf y},R_{\max}-r$}\;
  \If{$a=\texttt{STOP}$ \textbf{or} $r=R_{\max}$}{
    \Break\;
  }
  $\hat{\mathbf y}[S_j]\gets$ \Ttwo{$\mathbf x,\mathcal A,S_j,f$}\;
}
\tcp{T4: boundary reconciliation}
$(a,\mathit{patch})\gets$ \Tfour{$\mathbf x,\mathcal A,\mathcal S,\hat{\mathbf y}$}\;
\If{$a=\texttt{REGENERATE}$}{
  $\hat{\mathbf y}\gets$ \ApplyPatch{$\hat{\mathbf y},\mathit{patch}$}\;
}
\KwRet{$\hat{\mathbf y}$}\;
\end{algorithm}

%% file: tables/pjm_event_overlap.tex
\begin{table}[!htbp]
\caption{WAPE by number of events ($L=96$, $H=96$ for GEFCom2014; $L=168$, $H=24$ for PJM and SPP).}
\label{tab:app-pjm-overlap}
\centering\fontsize{6.5}{7.8}\selectfont
\setlength{\tabcolsep}{1pt}
\renewcommand{\arraystretch}{1}
\begin{tabular*}{\columnwidth}{@{\extracolsep{\fill}}ll*{5}{c}@{}}
\toprule
Dataset & Event Condition & $N$ & Full & Direct & No text & $\Delta$ \\
\midrule
\multirow{2}{*}{GEFCom2014} & Single Event & 72 & 6.42 & 7.90 & 8.20 & -1.48 \\
 & Multiple Events & 54 & 7.61 & 9.50 & 10.80 & -1.89 \\
\midrule
\multirow{3}{*}{PJM} & No Event & 35 & 2.63 & 2.80 & 2.41 & -0.17 \\
 & Single Event & 45 & 3.12 & 3.42 & 3.57 & -0.30 \\
 & Multiple Events & 32 & 3.69 & 4.31 & 4.63 & -0.62 \\
\midrule
\multirow{2}{*}{SPP} & Single Event & 18 & 2.74 & 2.91 & 3.08 & -0.17 \\
 & Multiple Events & 8 & 3.11 & 3.43 & 3.70 & -0.32 \\
\bottomrule
\end{tabular*}
\par\vspace{3pt}
{\raggedright $N$: cases per group; Full: ElecLoad (Full); Direct: direct-only; No text: without event text. $\Delta$: Full minus Direct before rounding (percentage points).\par}
\end{table}